\documentclass[11pt,a4paper]{article}

\usepackage[hyperref]{acl2021}
\usepackage{times}
\usepackage{latexsym}

\usepackage{amsmath,amsfonts,bm}

\def\eqref#1{equation~\ref{#1}}
\def\1{\bm{1}}

\DeclareMathAlphabet{\mathsfit}{\encodingdefault}{\sfdefault}{m}{sl}
\SetMathAlphabet{\mathsfit}{bold}{\encodingdefault}{\sfdefault}{bx}{n}

\newcommand{\E}{\mathbb{E}}

\newcommand{\R}{\mathbb{R}}

\usepackage{amsmath}
\usepackage{booktabs}
\usepackage{tablefootnote}
\usepackage{multirow}
\usepackage{makecell}
\usepackage{graphicx}
\usepackage{subcaption}
\usepackage{tabu}

\usepackage{footnote}
\makesavenoteenv{table}
\makesavenoteenv{table*}
\makesavenoteenv{tabular}

\newcommand{\tabincell}[2]{\begin{tabular}{@{}#1@{}}#2\end{tabular}}

\usepackage{hyperref}
\makeatletter
\def\UrlAlphabet{%
    \do\a\do\b\do\c\do\d\do\e\do\f\do\g\do\h\do\i\do\j%
    \do\k\do\l\do\m\do\n\do\o\do\p\do\q\do\r\do\s\do\t%
    \do\u\do\v\do\w\do\x\do\y\do\z\do\A\do\B\do\C\do\D%
    \do\E\do\F\do\G\do\H\do\I\do\J\do\K\do\L\do\M\do\N%
    \do\O\do\P\do\Q\do\R\do\S\do\T\do\U\do\V\do\W\do\X%
    \do\Y\do\Z}
\def\UrlDigits{\do\1\do\2\do\3\do\4\do\5\do\6\do\7\do\8\do\9\do\0}
\g@addto@macro{\UrlBreaks}{\UrlOrds}
\g@addto@macro{\UrlBreaks}{\UrlAlphabet}
\g@addto@macro{\UrlBreaks}{\UrlDigits}
\makeatother

\usepackage{microtype}

\aclfinalcopy % Uncomment this line for the final submission
\def\aclpaperid{337} %  Enter the acl Paper ID here

\title{LayoutLMv2: Multi-modal Pre-training for Visually-rich\\ Document Understanding}

\author{
	Yang Xu$^{1}$\thanks{\, Equal contributions during internship at MSRA}\,\ ,
	Yiheng Xu$^{2\,*}$,
	Tengchao Lv$^{2\,*}$,
	Lei Cui$^{2}$,
	Furu Wei$^{2}$,
	Guoxin Wang$^{3}$, \\
	{\bf Yijuan Lu$^{3}$,
	Dinei Florencio$^{3}$,
	Cha Zhang$^{3}$,
	Wanxiang Che$^{1}$,
	Min Zhang$^{4}$,
	Lidong Zhou$^{2}$} \\
	$^{1}$Research Center for Social Computing and Information Retrieval, \\Harbin Institute of Technology \\
	$^{2}$Microsoft Research Asia \quad $^{3}$Microsoft Azure AI \quad $^{4}$Soochow University \\
	$^{1}$\texttt{\{yxu,car\}@ir.hit.edu.cn}, \\
	$^{2}$\texttt{\{v-yixu,v-telv,lecu,fuwei,lidongz\}@microsoft.com}, \\
	$^{3}$\texttt{\{guow,yijlu,dinei,chazhang\}@microsoft.com}
	$^{4}$\texttt{minzhang@suda.edu.cn}
}

\date{}

\begin{document}
\maketitle
\begin{abstract}

Pre-training of text and layout has proved effective in a variety of visually-rich document understanding tasks due to its effective model architecture and the advantage of large-scale unlabeled scanned/digital-born documents. We propose \textbf{LayoutLMv2} architecture with new pre-training tasks to model the interaction among text, layout, and image in a single multi-modal framework. Specifically, with a two-stream multi-modal Transformer encoder, LayoutLMv2 uses not only the existing masked visual-language modeling task but also the new text-image alignment and text-image matching tasks, which make it better capture the cross-modality interaction in the pre-training stage. Meanwhile, it also integrates a spatial-aware self-attention mechanism into the Transformer architecture so that the model can fully understand the relative positional relationship among different text blocks. Experiment results show that LayoutLMv2 outperforms LayoutLM by a large margin and achieves new state-of-the-art results on a wide variety of downstream visually-rich document understanding tasks, including FUNSD (0.7895 $\to$ 0.8420), CORD (0.9493 $\to$ 0.9601), SROIE (0.9524 $\to$ 0.9781), Kleister-NDA (0.8340 $\to$ 0.8520), RVL-CDIP (0.9443 $\to$ 0.9564), and DocVQA (0.7295 $\to$ 0.8672). We made our model and code publicly available at \url{https://aka.ms/layoutlmv2}.
\end{abstract}

\section{Introduction}

Visually-rich Document Understanding (VrDU) aims to analyze scanned/digital-born business documents (images of invoices, forms in PDF format, etc.) where structured information can be automatically extracted and organized for many business applications. Distinct from conventional information extraction tasks, the VrDU task relies on not only textual information but also visual and layout information that is vital for visually-rich documents.
Different types of documents indicate that the text fields of interest located at different positions within the document, which is often determined by the style and format of each type as well as the document content. Therefore, to accurately recognize the text fields of interest, it is inevitable to take advantage of the cross-modality nature of visually-rich documents, where the textual, visual, and layout information should be jointly modeled and learned end-to-end in a single framework.

The recent progress of VrDU lies primarily in two directions. The first direction is usually built on the shallow fusion between textual and visual/layout/style information~\citep{Yang_2017,liu-etal-2019-graph,ijcai2019-466,yu2020pick,majumder-etal-2020-representation,Wei_2020,zhang2020trie}. These approaches leverage the pre-trained NLP and CV models individually and combine the information from multiple modalities for supervised learning.
Although good performance has been achieved, the domain knowledge of one document type cannot be easily transferred into another, so that these models often need to be re-trained once the document type is changed. Thereby the local invariance in general document layout (key-value pairs in a left-right layout, tables in a grid layout, etc.) cannot be fully exploited.
To this end, the second direction relies on the deep fusion among textual, visual, and layout information from a great number of unlabeled documents in different domains, where pre-training techniques play an important role in learning the cross-modality interaction in an end-to-end fashion~\citep{Lockard_2020,10.1145/3394486.3403172}.
In this way, the pre-trained models absorb cross-modal knowledge from different document types, where the local invariance among these layouts and styles is preserved. Furthermore, when the model needs to be transferred into another domain with different document formats, only a few labeled samples would be sufficient to fine-tune the generic model in order to achieve state-of-the-art accuracy.
Therefore, the proposed model in this paper follows the second direction, and we explore how to further improve the pre-training strategies for the VrDU tasks.

In this paper, we present an improved version of LayoutLM~\citep{10.1145/3394486.3403172}, aka \textbf{LayoutLMv2}.
Different from the vanilla LayoutLM model where visual embeddings are combined in the fine-tuning stage, we integrate the visual information in the pre-training stage in LayoutLMv2 by taking advantage of the Transformer architecture to learn the cross-modality interaction between visual and textual information.
In addition, inspired by the 1-D relative position representations~\citep{Shaw_2018, raffel2020exploring, bao2020unilmv2}, we propose the spatial-aware self-attention mechanism for LayoutLMv2, which involves a 2-D relative position representation for token pairs. Different from the absolute 2-D position embeddings that LayoutLM uses to model the page layout, the relative position embeddings explicitly provide a broader view for the contextual spatial modeling.
For the pre-training strategies, we use two new training objectives for LayoutLMv2 in addition to the masked visual-language modeling.
The first is the proposed text-image alignment strategy, which aligns the text lines and the corresponding image regions.
The second is the text-image matching strategy popular in previous vision-language pre-training models~\citep{tan2019lxmert,lu2019vilbert,su2020vlbert,chen2020uniter,Sun_2019_ICCV}, where the model learns whether the document image and textual content are correlated.

We select six publicly available benchmark datasets as the downstream tasks to evaluate the performance of the pre-trained LayoutLMv2 model, which are the FUNSD dataset~\citep{Jaume_2019} for form understanding, the CORD dataset~\citep{park2019cord} and the SROIE dataset~\citep{8977955} for receipt understanding, \iffalse \footnote{\url{https://rrc.cvc.uab.es/?ch=13}} \fi the Kleister-NDA dataset~\citep{graliski2020kleister} for long document understanding with a complex layout, the RVL-CDIP dataset~\citep{harley2015icdar} for document image classification, and the DocVQA dataset~\citep{Mathew_2021_WACV} for visual question answering on document images. Experiment results show that the LayoutLMv2 model significantly outperforms strong baselines, including the vanilla LayoutLM, and achieves new state-of-the-art results in all of these tasks.

\begin{figure*}[t]
    \centering
    \includegraphics[width=0.8\textwidth]{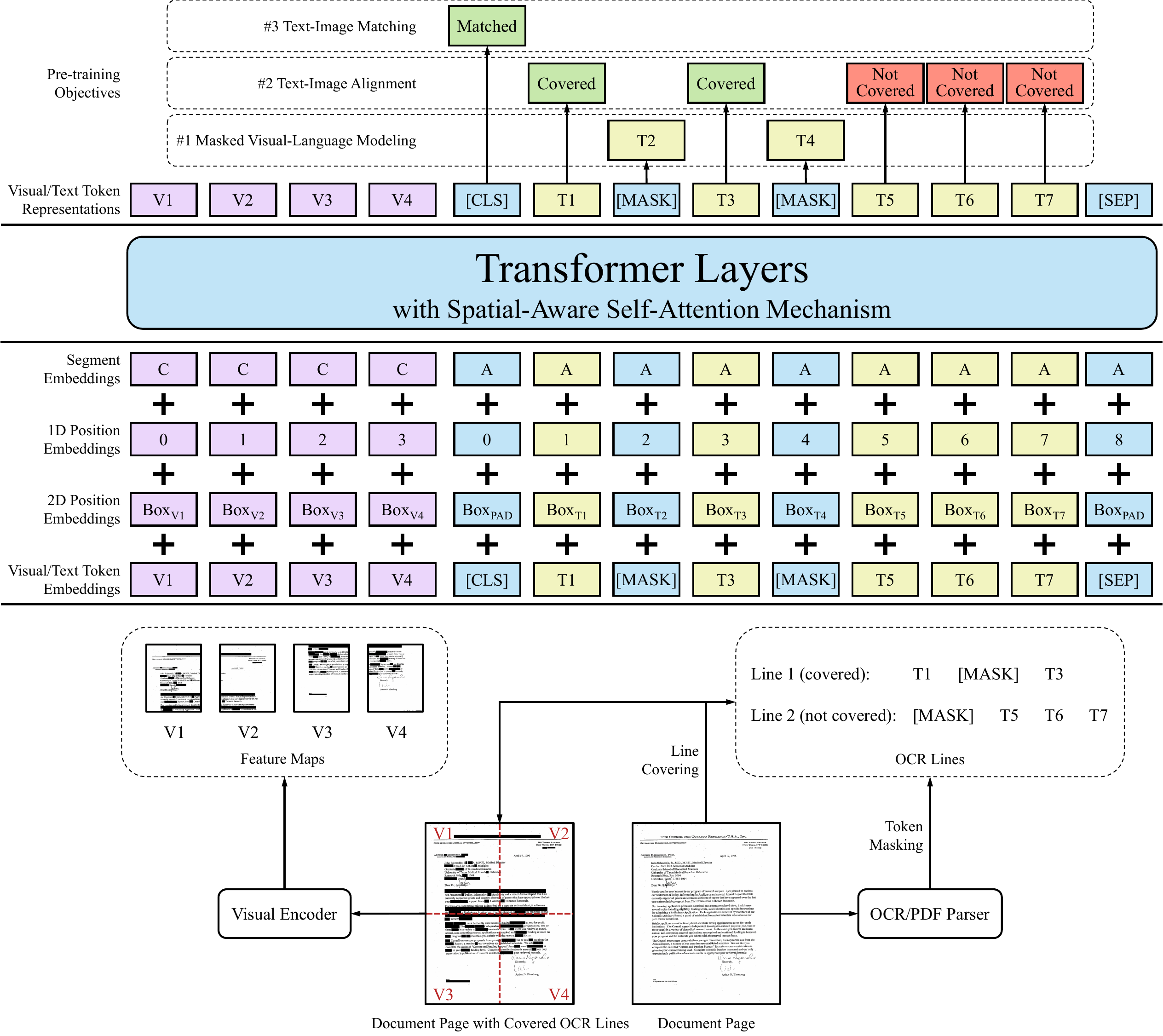}
    \caption{An illustration of the model architecture and pre-training strategies for LayoutLMv2}
    \label{fig:2.LayoutLMv2}
\end{figure*}

The contributions of this paper are summarized as follows:

\begin{itemize}

    \item We propose a multi-modal Transformer model to integrate the document text, layout, and visual information in the pre-training stage, which learns the cross-modal interaction end-to-end in a single framework. Meanwhile, a spatial-aware self-attention mechanism is integrated into the Transformer architecture.
    \item In addition to the masked visual-language model, we add text-image alignment and text-image matching as the new pre-training strategies to enforce the alignment among different modalities.
    \item LayoutLMv2 significantly outperforms and achieves new SOTA results not only on the conventional VrDU tasks but also on the VQA task for document images, which demonstrates the great potential for the multi-modal pre-training for VrDU.
\end{itemize}

\section{Approach}

In this section, we will introduce the model architecture and the multi-modal pre-training tasks of LayoutLMv2, which is illustrated in Figure \ref{fig:2.LayoutLMv2}.

\subsection{Model Architecture}

We build a multi-modal Transformer architecture as the backbone of LayoutLMv2, which takes text, visual, and layout information as input to establish deep cross-modal interactions. We also introduce a spatial-aware self-attention mechanism to the model architecture for better modeling the document layout.
Detailed descriptions of the model are as follows.

\paragraph{Text Embedding}

Following the common practice, we use WordPiece \citep{wu2016google} to tokenize the OCR text sequence and assign each token to a certain segment $s_i \in \{\texttt{[A]}, \texttt{[B]}\}$. Then, we add {\tt[CLS]} at the beginning of the sequence and {\tt[SEP]} at the end of each text segment. Extra {\tt[PAD]} tokens are appended to the end so that the final sequence's length is exactly the maximum sequence length $L$.
The final text embedding is the sum of three embeddings. Token embedding represents the token itself, 1D positional embedding represents the token index, and segment embedding is used to distinguish different text segments. Formally, we have the $i$-th ($0 \le i < L$) text embedding
$$\mathbf{t}_i = \mathrm{TokEmb}(w_i) + \mathrm{PosEmb1D}(i) + \mathrm{SegEmb}(s_i)$$

\paragraph{Visual Embedding}

Although all information we need is contained in the page image, the model has difficulty capturing detailed features in a single information-rich representation of the entire page. Therefore, we leverage the output feature map of a CNN-based visual encoder, which converts the page image to a fixed-length sequence.
We use ResNeXt-FPN~\citep{Xie2016AggregatedRT,Lin_2017_CVPR} architecture as the backbone of the visual encoder, whose parameters can be updated through backpropagation.

Given a document page image $I$, it is resized to $224 \times 224$ then fed into the visual backbone.
After that, the output feature map is average-pooled to a fixed size with the width being $W$ and height being $H$. Next, it is flattened into a visual embedding sequence of length $W \times H$. The sequence is named $\mathrm{VisTokEmb}(I)$.
A linear projection layer is then applied to each visual token embedding to unify the dimensionality with the text embeddings.
Since the CNN-based visual backbone cannot capture the positional information, we also add a 1D positional embedding to these visual token embeddings. The 1D positional embedding is shared with the text embedding layer.
For the segment embedding, we attach all visual tokens to the visual segment {\tt[C]}.
The $i$-th ($0 \le i < WH$) visual embedding can be represented as
\begin{equation*}
    \begin{split}
        \mathbf{v}_i = & \, \mathrm{Proj}\bigl(\mathrm{VisTokEmb}(I)_i\bigr) \\
                       & \!+\! \mathrm{PosEmb1D}(i) \!+\! \mathrm{SegEmb}(\texttt{[C]})
    \end{split}
\end{equation*}

\paragraph{Layout Embedding}

The layout embedding layer is for embedding the spatial layout information represented by axis-aligned token bounding boxes from the OCR results, in which box width and height together with corner coordinates are identified. Following the vanilla LayoutLM, we normalize and discretize all coordinates to integers in the range $[0, 1000]$, and use two embedding layers to embed $\mathrm{x}$-axis features and $\mathrm{y}$-axis features separately. Given the normalized bounding box of the $i$-th ($0 \le i < WH + L$) text/visual token $\mathrm{box}_i = (x_{\rm min}, x_{\rm max}, y_{\rm min}, y_{\rm max}, width, height)$, the layout embedding layer concatenates six bounding box features to construct a token-level 2D positional embedding, aka the layout embedding

\begin{equation*}
    \begin{split}
        \mathbf{l}_i = \mathrm{Concat}\bigl( & \mathrm{PosEmb2D_x}(x_{\rm min}, x_{\rm max}, width), \\
                                             & \mathrm{PosEmb2D_y}(y_{\rm min}, y_{\rm max}, height)\bigr)
    \end{split}
\end{equation*}

Note that CNNs perform local transformation, thus the visual token embeddings can be mapped back to image regions one by one with neither overlap nor omission. When calculating bounding boxes, the visual tokens can be treated as evenly divided grids. An empty bounding box $\mathrm{box}_{\mathrm{PAD}} = (0, 0, 0, 0, 0, 0)$ is attached to special tokens {\tt[CLS]}, {\tt[SEP]} and {\tt[PAD]}.

\paragraph{Multi-modal Encoder with Spatial-Aware Self-Attention Mechanism}
The encoder concatenates visual embeddings $\{\mathbf{v}_0, ..., \mathbf{v}_{WH - 1}\}$ and text embeddings $\{\mathbf{t}_0, ..., \mathbf{t}_{L - 1}\}$ to a unified sequence
and fuses spatial information by adding the layout embeddings to get the $i$-th ($0 \le i < WH + L$) first layer input
\begin{equation*}
    \begin{split}
        \mathbf{x}^{(0)}_i & = X_i + \mathbf{l}_i,\, \text{where} \\
        X & = \{\mathbf{v}_0, ..., \mathbf{v}_{WH - 1}, \mathbf{t}_0, ..., \mathbf{t}_{L - 1}\}
    \end{split}
\end{equation*}

Following the architecture of Transformer, we build our multi-modal encoder with a stack of multi-head self-attention layers followed by a feed-forward network. However, the original self-attention mechanism can only implicitly capture the relationship between the input tokens with the absolute position hints. In order to efficiently model local invariance in the document layout, it is necessary to insert relative position information explicitly. Therefore, we introduce the spatial-aware self-attention mechanism into the self-attention layers. For simplicity, the following description is for a single head in a single self-attention layer with hidden size of $d_{head}$ and projection matrics $\mathbf{W}^Q$, $\mathbf{W}^K$, $\mathbf{W}^V$. The original self-attention mechanism captures the correlation between query $\mathbf{x}_i$ and key $\mathbf{x}_j$ by projecting the two vectors and calculating the attention score
$$
\mathbf{\alpha}_{i j} =
\frac{1}{\sqrt{d_{head}}}
\left(\mathbf{x}_i\mathbf{W}^Q\right)
\left(\mathbf{x}_j\mathbf{W}^K\right)^\mathsf{T}
$$
Considering the large range of positions, we model the semantic relative position and spatial relative position as bias terms to prevent adding too many parameters.
Similar practice has been shown effective on text-only Transformer architectures~\citep{raffel2020exploring, bao2020unilmv2}.
Let $\mathbf{b}^{\mathrm{(1D)}}$, $\mathbf{b}^{\mathrm{(2D_x)}}$ and $\mathbf{b}^{\mathrm{(2D_y)}}$ denote the learnable 1D and 2D relative position biases respectively. The biases are different among attention heads but shared in all encoder layers. Assuming $(x_i, y_i)$ anchors the top left corner coordinates of the $i$-th bounding box, we obtain the spatial-aware attention score
$$
\mathbf{\alpha}_{i j}^{\prime} =
\mathbf{\alpha}_{i j}
+ \mathbf{b}^{\mathrm{(1D)}}_{j - i}
+ \mathbf{b}^{\mathrm{(2D_x)}}_{x_j - x_i}
+ \mathbf{b}^{\mathrm{(2D_y)}}_{y_j - y_i}
$$

Finally, the output vectors are represented as the weighted average of all the projected value vectors with respect to normalized spatial-aware attention scores
$$
\mathbf{h}_{i} =
\sum_{j}
\frac{
    \exp \left(\alpha'_{i j}\right)}
{\sum_{k} \exp \left(\alpha'_{i k}\right)}\mathbf{x}_j\mathbf{W}^V
$$

\subsection{Pre-training Tasks}

\paragraph{Masked Visual-Language Modeling}

Similar to the vanilla LayoutLM, we use the Masked Visual-Language Modeling (MVLM) to make the model learn better in the language side with the cross-modality clues. We randomly mask some text tokens and ask the model to recover the masked tokens. Meanwhile, the layout information remains unchanged, which means the model knows each masked token's location on the page. The output representations of masked tokens from the encoder are fed into a classifier over the whole vocabulary, driven by a cross-entropy loss.
To avoid visual clue leakage, we mask image regions corresponding to masked tokens on the raw page image input before feeding it into the visual encoder.

\paragraph{Text-Image Alignment}

To help the model learn the spatial location correspondence between image and coordinates of bounding boxes, we propose the Text-Image Alignment (TIA) as a fine-grained cross-modality alignment task. In the TIA task, some tokens lines are randomly selected, and their image regions are covered on the document image. We call this operation covering to avoid confusion with the masking operation in MVLM. During pre-training, a classification layer is built above the encoder outputs. This layer predicts a label for each text token depending on whether it is covered, i.e., {\tt[Covered]} or {\tt[Not Covered]}, and computes the binary cross-entropy loss.
Considering the input image's resolution is limited, and some document elements like signs and bars in a figure may look like covered text regions, the task of finding a word-sized covered image region can be noisy. Thus, the covering operation is performed at the line-level.
When MVLM and TIA are performed simultaneously, TIA losses of the tokens masked in MVLM are not taken into account. This prevents the model from learning the useless but straightforward correspondence from {\tt[MASK]} to {\tt[Covered]}.

\paragraph{Text-Image Matching}

Furthermore, a coarse-grained cross-modality alignment task, Text-Image Matching (TIM) is applied to help the model learn the correspondence between document image and textual content. We feed the output representation at {\tt[CLS]} into a classifier to predict whether the image and text are from the same document page. Regular inputs are positive samples. To construct a negative sample, an image is either replaced by a page image from another document or dropped.
To prevent the model from cheating by finding task features, we perform the same masking and covering operations to images in negative samples. The TIA target labels are all set to {\tt[Covered]} in negative samples. We apply the binary cross-entropy loss in the optimization process.

\section{Experiments}

\subsection{Data}

In order to pre-train and evaluate LayoutLMv2 models, we select datasets in a wide range from the visually-rich document understanding area. Following LayoutLM, we use IIT-CDIP Test Collection~\citep{10.1145/1148170.1148307} as the pre-training dataset. Six datasets are used as down-stream tasks. The FUNSD~\citep{Jaume_2019}, CORD~\citep{park2019cord}, SROIE~\citep{8977955} and Kleister-NDA~\citep{graliski2020kleister} datasets define entity extraction tasks that aim to extract the value of a set of pre-defined keys, which we formalize as a sequential labeling task. RVL-CDIP~\citep{harley2015icdar} is for document image classification. DocVQA~\citep{Mathew_2021_WACV}, as the name suggests, is a dataset for visual question answering on document images. Statistics of datasets are shown in Table \ref{tab:data}. Refer to the Appendix for details.

\begin{table}[ht]
    \centering
    \small
    \begin{tabular}{lll}
        \toprule
        \multicolumn{1}{c}{\bf Dataset} & \bf \tabincell{c}{\# of keys or\\categories} & \bf \tabincell{c}{\# of examples\\(train/dev/test)}  \\
        \midrule
        IIT-CDIP & -- & 11M/0/0  \\
        FUNSD & 4 & 149/0/50  \\
        CORD & 30 & 800/100/100  \\
        SROIE & 4 & 626/0/347  \\
        Kleister-NDA & 4 & 254/83/203  \\
        RVL-CDIP & 16 & 320K/4K/4K  \\
        DocVQA & -- & 39K/5K/5K  \\
        \bottomrule
    \end{tabular}
    \caption{Statistics of datasets}
    \label{tab:data}
\end{table}

\subsection{Settings}
\label{sec:Experiments.Settings}

Following the typical pre-training and fine-tuning strategy, we update all parameters including the visual encoder layers, and train whole models end-to-end for all the settings. Training details can be found in the Appendix.

\begin{table*}[ht]
    \centering
    \small
    \begin{tabular}{lcccc}
        \toprule
        \multicolumn{1}{c}{\bf Model} & \bf FUNSD & \bf CORD & \bf SROIE & \bf Kleister-NDA \\
        \midrule
        $\textrm{BERT}_{\rm BASE}$ & 0.6026 & 0.8968 & 0.9099 & 0.7790  \\
        $\textrm{UniLMv2}_{\rm BASE}$ & 0.6890 & 0.9092 & 0.9459 & 0.7950  \\
        $\textrm{BERT}_{\rm LARGE}$ & 0.6563 & 0.9025 & 0.9200 & 0.7910  \\
        $\textrm{UniLMv2}_{\rm LARGE}$ & 0.7257 & 0.9205 & 0.9488 & 0.8180  \\
        \midrule
        $\textrm{LayoutLM}_{\rm BASE}$ & 0.7866 & 0.9472 & 0.9438 & 0.8270  \\
        $\textrm{LayoutLM}_{\rm LARGE}$ & 0.7895 & 0.9493 & 0.9524 & 0.8340  \\
        \midrule
        $\textrm{LayoutLMv2}_{\rm BASE}$ & 0.8276 & 0.9495 & 0.9625 & 0.8330  \\
        $\textrm{LayoutLMv2}_{\rm LARGE}$ & \bf 0.8420 & \bf 0.9601 & \bf 0.9781 & \bf 0.8520  \\
        \midrule
        \midrule
        BROS~\citep{hong2021bros} & 0.8121 & 0.9536 & 0.9548 & -- \\
        SPADE~\citep{hwang2020spatial} & -- & 0.9150 & -- & --  \\
        PICK~\citep{yu2020pick} & -- & -- & 0.9612 & --  \\
        TRIE~\citep{zhang2020trie} & -- & -- & 0.9618 & --  \\
        Top-1 on SROIE Leaderboard (until 2020-12-24) & -- & -- & 0.9767 & --  \\
        $\textrm{RoBERTa}_{\rm BASE}$ in~\citep{graliski2020kleister} & -- & -- & -- & 0.7930  \\
        \bottomrule
    \end{tabular}
    \caption{Entity-level F1 scores of the four entity extraction tasks: FUNSD, CORD, SROIE and Kleister-NDA. Detailed per-task results are in the Appendix.}
    \label{tab:kie}
\end{table*}

\paragraph{Pre-training LayoutLMv2}

We train LayoutLMv2 models with two different parameter sizes. We use a 12-layer 12-head Transformer encoder and set hidden size $d = 768$ in LayoutLMv2$_{\rm BASE}$. While in the LayoutLMv2$_{\rm LARGE}$, the encoder has 24 Transformer layers with 16 heads and $d = 1024$. Visual backbones in the two models are based on the same ResNeXt101-FPN architecture. The numbers of parameters are 200M and 426M approximately for LayoutLMv2$_{\rm BASE}$ and LayoutLMv2$_{\rm LARGE}$, respectively.

For the encoder along with the text embedding layer, LayoutLMv2 uses the same architecture as UniLMv2~\citep{bao2020unilmv2}, thus it is initialized from UniLMv2. For the ResNeXt-FPN part in the visual embedding layer, the backbone of a Mask-RCNN~\citep{He_2017_ICCV} model trained on PubLayNet \citep{zhong2019publaynet} is leveraged.\footnote{``MaskRCNN ResNeXt101\_32x8d FPN 3X" setting in \url{https://github.com/hpanwar08/detectron2}} The rest of the parameters in the model are randomly initialized.

During pre-training, we sample pages from the IIT-CDIP dataset and select a random sliding window of the text sequence if the sample is too long. We set the maximum sequence length $L = 512$ and assign all text tokens to the segment {\tt[A]}. The output shape of the average pooling layer is set to $W = H = 7$, so that it transforms the feature map into $49$ visual tokens. In MVLM, 15\% text tokens are masked among which 80\% are replaced by a special token {\tt[MASK]}, 10\% are replaced by a random token sampled from the whole vocabulary, and 10\% remains the same. In TIA, 15\% of the lines are covered. In TIM, 15\% images are replaced, and 5\% are dropped.

\paragraph{Fine-tuning LayoutLMv2}

We use the {\tt[CLS]} output along with pooled visual token representations as global features in the document-level classification task RVL-CDIP.
For the extractive question answering task DocVQA and the other four entity extraction tasks, we follow common practice like ~\cite{devlin-etal-2019-bert} and build task specified head layers over the text part of LayoutLMv2 outputs.

In the DocVQA paper, experiment results show that the BERT model fine-tuned on the SQuAD dataset~\citep{rajpurkar-etal-2016-squad} outperforms the original BERT model. Inspired by this fact, we add an extra setting, which is that we first fine-tune LayoutLMv2 on a question generation (QG) dataset followed by the DocVQA dataset. The QG dataset contains almost one million question-answer pairs generated by a generation model trained on the SQuAD dataset.

\paragraph{Baselines}

We select three baseline models in the experiments to compare LayoutLMv2 with the text-only pre-trained models as well as the vanilla LayoutLM model. Specifically, we compare LayoutLMv2 with BERT~\citep{devlin-etal-2019-bert}, UniLMv2~\citep{bao2020unilmv2}, and LayoutLM~\citep{10.1145/3394486.3403172} for all the experiment settings. We use the publicly available PyTorch models for BERT~\citep{wolf-etal-2020-transformers} and LayoutLM, and use our in-house implementation for the UniLMv2 models. For each baseline approach, experiments are conducted using both the $\mathrm{BASE}$ and $\mathrm{LARGE}$ parameter settings.

\subsection{Results}

\paragraph{Entity Extraction Tasks}

Table~\ref{tab:kie} shows the model accuracy on the four datasets FUNSD, CORD, SROIE, and Kleister-NDA, which we regard as sequential labeling tasks evaluated using entity-level F1 score. We report the evaluation results of Kleister-NDA on the validation set because the ground-truth labels and the submission website for the test set are not available right now.
For text-only models, the UniLMv2 models outperform the BERT models by a large margin in terms of the $\mathrm{BASE}$ and $\mathrm{LARGE}$ settings. For text+layout models, the LayoutLM family, especially the LayoutLMv2 models, brings significant performance improvement over the text-only baselines.
Compared to the baselines, the LayoutLMv2 models are superior to the SPADE~\citep{hwang2020spatial} decoder method, as well as the text+layout pre-training approach BROS~\citep{hong2021bros} that is built on the SPADE decoder, which demonstrates the effectiveness of our modeling approach. Moreover, with the same modal information, our LayoutLMv2 models also outperform existing multi-modal approaches PICK~\citep{yu2020pick}, TRIE~\citep{zhang2020trie} and the previous top-1 method on the leaderboard,\footnote{Unpublished results, the leaderboard is available at  \url{https://rrc.cvc.uab.es/?ch=13&com=evaluation&task=3}} confirming the effectiveness of our pre-training for text, layout, and visual information.
The best performance on all the four datasets is achieved by the $\textrm{LayoutLMv2}_{\rm LARGE}$, which illustrates that the multi-modal pre-training in LayoutLMv2 learns better from the interactions from different modalities, thereby leading to the new SOTA on various document understanding tasks.

\begin{table}[t]
    \centering
    \small
    \begin{tabular}{lc}
        \toprule
        \multicolumn{1}{c}{\bf Model} & \bf Accuracy   \\\midrule
        $\textrm{BERT}_{\rm BASE}$  &  89.81\% \\
        $\textrm{UniLMv2}_{\rm BASE}$  &  90.06\% \\
        $\textrm{BERT}_{\rm LARGE}$   & 89.92\%    \\
        $\textrm{UniLMv2}_{\rm LARGE}$  &  90.20\% \\\midrule
        $\textrm{LayoutLM}_{\rm BASE}$ (w/ image) & 94.42\%  \\
        $\textrm{LayoutLM}_{\rm LARGE}$ (w/ image) & 94.43\%  \\
        \midrule
        $\textrm{LayoutLMv2}_{\rm BASE}$ & 95.25\%  \\
        $\textrm{LayoutLMv2}_{\rm LARGE}$ & \bf 95.64\%  \\
        \midrule\midrule
        VGG-16~\citep{Afzal2017CuttingTE} & 90.97\%  \\
        Single model~\citep{Das2018DocumentIC} & 91.11\% \\
        Ensemble~\citep{Das2018DocumentIC} & 92.21\% \\
        InceptionResNetV2~\citep{Szegedy2016Inceptionv4IA} & 92.63\% \\
        LadderNet~\citep{ijcai2019-466} & 92.77\% \\
        Single model~\citep{Dauphinee2019ModularMA} &  93.03\% \\
        Ensemble~\citep{Dauphinee2019ModularMA} &  93.07\% \\
        \bottomrule
    \end{tabular}
    \caption{Classification accuracy on the RVL-CDIP dataset}
    \label{tab:rvlcdip}
\end{table}

\paragraph{RVL-CDIP}

Table~\ref{tab:rvlcdip} shows the classification accuracy on the RVL-CDIP dataset, including text-only pre-trained models, the LayoutLM family as well as several image-based baseline models. As shown in the table, both the text and visual information are important to the document image classification task because document images are text-intensive and represented by a variety of layouts and formats. Therefore, we observed that the LayoutLM family outperforms those text-only or image-only models as it leverages the multi-modal information within the documents. Specifically, the $\textrm{LayoutLMv2}_{\rm LARGE}$ model significantly improves the classification accuracy by more than 1.2\% point over the previous SOTA results, which achieves an accuracy of 95.64\%. This also verifies that the pre-trained LayoutLMv2 model benefits not only the information extraction tasks in  document understanding but also the document image classification task through effective multi-model training.

\begin{table}[t]
    \centering
    \small
    \begin{tabular}{llc}
        \toprule
        \multicolumn{1}{c}{\bf Model} & \bf Fine-tuning set & \bf ANLS   \\
        \midrule
        $\textrm{BERT}_{\rm BASE}$ & train & 0.6354  \\
        $\textrm{UniLMv2}_{\rm BASE}$ & train & 0.7134  \\
        $\textrm{BERT}_{\rm LARGE}$  & train & 0.6768  \\
        $\textrm{UniLMv2}_{\rm LARGE}$ & train & 0.7709  \\
        \midrule
        $\textrm{LayoutLM}_{\rm BASE}$ & train & 0.6979  \\
        $\textrm{LayoutLM}_{\rm LARGE}$ & train & 0.7259  \\
        \midrule
        $\textrm{LayoutLMv2}_{\rm BASE}$ & train & 0.7808  \\
        $\textrm{LayoutLMv2}_{\rm LARGE}$ & train & 0.8348  \\
        \midrule
        $\textrm{LayoutLMv2}_{\rm LARGE}$ & train + dev & 0.8529  \\

        $\textrm{LayoutLMv2}_{\rm LARGE}$ + QG & train + dev & \bf 0.8672  \\
        \midrule\midrule
        \tabincell{l}{Top-1 (30 models ensemble)\\on DocVQA Leaderboard\\(until 2020-12-24)} & - & 0.8506  \\
        \bottomrule
    \end{tabular}
    \caption{ANLS score on the DocVQA dataset, ``QG" denotes the data augmentation with the question generation dataset.}
    \label{tab:docvqa}
\end{table}

\begin{table*}[ht]
    \centering
    \small
    \begin{tabular}{cllccccc}
        \toprule
        \bf \# & \bf Model Architecture & \bf Initialization & \bf SASAM & \bf MVLM & \bf TIA & \bf TIM & \bf ANLS \\
        \midrule
        1 & $\textrm{LayoutLM}_{\rm BASE}$ & $\textrm{BERT}_{\rm BASE}$ & & \checkmark & & & 0.6841 \\
        \midrule
        2a & $\textrm{LayoutLMv2}_{\rm BASE}$ & $\textrm{BERT}_{\rm BASE}$ + X101-FPN & & \checkmark & & & 0.6915 \\
        2b & $\textrm{LayoutLMv2}_{\rm BASE}$ & $\textrm{BERT}_{\rm BASE}$ + X101-FPN & & \checkmark & \checkmark & & 0.7061 \\
        2c & $\textrm{LayoutLMv2}_{\rm BASE}$ & $\textrm{BERT}_{\rm BASE}$ + X101-FPN & & \checkmark & & \checkmark & 0.6955 \\
        2d & $\textrm{LayoutLMv2}_{\rm BASE}$ & $\textrm{BERT}_{\rm BASE}$ + X101-FPN & & \checkmark & \checkmark & \checkmark & 0.7124 \\
        \midrule
        3 & $\textrm{LayoutLMv2}_{\rm BASE}$ & $\textrm{BERT}_{\rm BASE}$ + X101-FPN & \checkmark & \checkmark & \checkmark & \checkmark & 0.7217 \\
        \midrule
        4 & $\textrm{LayoutLMv2}_{\rm BASE}$ & $\textrm{UniLMv2}_{\rm BASE}$ + X101-FPN & \checkmark & \checkmark & \checkmark & \checkmark & 0.7421 \\
        \bottomrule
    \end{tabular}
    \caption{Ablation study on the DocVQA dataset, where ANLS scores on the validation set are reported. ``SASAM" means the spatial-aware self-attention mechanism. ``MVLM", ``TIA" and ``TIM" are the three pre-training tasks. All the models are trained using the whole pre-training dataset for one epoch with the $\mathrm{BASE}$ model size.}
    \label{tab:ablation-docvqa-bert}
\end{table*}

\paragraph{DocVQA}

Table~\ref{tab:docvqa} lists the Average Normalized Levenshtein Similarity (ANLS) scores on the DocVQA dataset of text-only baselines, LayoutLM family models, and the previous top-1 on the leaderboard. With multi-modal pre-training, LayoutLMv2 models outperform LayoutLM models and text-only baselines by a large margin when fine-tuned on the train set. By using all data (train + dev) as the fine-tuning dataset, the $\textrm{LayoutLMv2}_{\rm LARGE}$ single model outperforms the previous top-1 on the leaderboard which ensembles 30 models.\footnote{Unpublished results, the leaderboard is available at \url{https://rrc.cvc.uab.es/?ch=17&com=evaluation&task=1}} Under the setting of fine-tuning $\textrm{LayoutLMv2}_{\rm LARGE}$ on a question generation dataset (QG) and the DocVQA dataset successively, the single model performance increases by more than 1.6\% ANLS and achieves the new SOTA.

\subsection{Ablation Studies}

To fully understand the underlying impact of different components, we conduct an ablation study to explore the effect of visual information, the pre-training tasks, spatial-aware self-attention mechanism, as well as different text-side initialization models. Table~\ref{tab:ablation-docvqa-bert} shows model performance on the DocVQA validation set. Under all the settings, we pre-train the models using all IIT-CDIP data for one epoch. The hyper-parameters are the same as those used to pre-train $\textrm{LayoutLMv2}_{\rm BASE}$ in Section~\ref{sec:Experiments.Settings}. ``LayoutLM" denotes the vanilla LayoutLM architecture in~\citep{10.1145/3394486.3403172}, which can be regarded as a LayoutLMv2 architecture without visual module and spatial-aware self-attention mechanism. ``X101-FPN" denotes the ResNeXt101-FPN visual backbone described in Section~\ref{sec:Experiments.Settings}.

We first evaluate the effect of introducing visual information. From \#1 to \#2a, we add the visual module without changing the pre-training strategy, where results show that LayoutLMv2 pre-trained with only MVLM can leverage visual information effectively. Then, we compare the two cross-modality alignment pre-training tasks TIA and TIM. According to the four results in \#2, both tasks improve the model performance substantially, and the proposed TIA benefits the model more than the commonly used TIM. Using both tasks together is more effective than using either one alone. According to this observation, we keep all the three pre-training tasks and introduce the spatial-aware self-attention mechanism (SASAM) to the model architecture. Compare the results \#2d and \#3, the proposed SASAM can further improve the model accuracy. Finally, in settings \#3 and \#4, we change the text-side initialization checkpoint from BERT to UniLMv2, and confirm that LayoutLMv2 benefits from the better initialization.

\section{Related Work}
In recent years, pre-training techniques have become popular in both NLP and CV areas, and have also been leveraged in the VrDU tasks.

~\citet{devlin-etal-2019-bert} introduced a new language representation model called BERT, which is designed to pre-train deep bidirectional representations from the unlabeled text by jointly conditioning on both left and right context in all layers.
~\citet{bao2020unilmv2} propose to pre-train a unified language model for both autoencoding and partially autoregressive language modeling tasks using a novel training procedure, referred to as a pseudo-masked language model.
Our multi-modal Transformer architecture and the MVLM pre-training strategy extend Transformer and MLM used in these work to leverage visual information.

~\citet{lu2019vilbert} proposed ViLBERT for learning task-agnostic joint representations of image content and natural language by extending the popular BERT architecture to a multi-modal two-stream model.
~\citet{su2020vlbert} proposed VL-BERT that adopts the Transformer model as the backbone, and extends it to take both visual and linguistic embedded features as input.
Different from these vision-language pre-training approaches, the visual part of LayoutLMv2 directly uses the feature map instead of pooled ROI features, and benefits from the new TIA pre-training task.

~\citet{10.1145/3394486.3403172} proposed LayoutLM to jointly model interactions between text and layout information across scanned document images, benefiting a great number of real-world document image understanding tasks such as information extraction from scanned documents.
This work is a natural extension of the vanilla LayoutLM, which takes advantage of textual, layout, and visual information in a single multi-modal pre-training framework.

\section{Conclusion}

In this paper, we present a multi-modal pre-training approach for visually-rich document understanding tasks, aka LayoutLMv2. Distinct from existing methods for VrDU, the LayoutLMv2 model not only considers the text and layout information but also integrates the image information in the pre-training stage with a single multi-modal framework. Meanwhile, the spatial-aware self-attention mechanism is integrated into the
Transformer architecture to capture the relative relationship among different bounding boxes. Furthermore, new pre-training objectives are also leveraged to enforce the learning of cross-modal interaction among different modalities. Experiment results on 6 different VrDU tasks have illustrated that the pre-trained LayoutLMv2 model has substantially outperformed the SOTA baselines in the document intelligence area, which greatly benefits a number of real-world document understanding tasks.

For future research, we will further explore the network architecture as well as the pre-training strategies for the LayoutLM family. Meanwhile, we will also investigate the language expansion to make the multi-lingual LayoutLMv2 model available for different languages, especially the non-English areas around the world.

\section*{Acknowledgments}

This work was supported by the National Key R\&D Program of China via grant 2020AAA0106501 and the National Natural Science Foundation of China (NSFC) via grant 61976072 and 61772153.

\bibliographystyle{acl_natbib}
\bibliography{anthology,acl2021}

\appendix

\section*{Appendix}

\section{Details of Datasets}

Introduction to the dataset and task definitions along with the description of required data processing are presented as follows.

\paragraph{Pre-training Dataset}

Following LayoutLM, we pre-train LayoutLMv2 on the IIT-CDIP Test Collection~\citep{10.1145/1148170.1148307}, which contains over 11 million scanned document pages. We extract text and corresponding word-level bounding boxes from document page images with the Microsoft Read API.\footnote{\url{https://docs.microsoft.com/en-us/azure/cognitive-services/computer-vision/concept-recognizing-text}}

\paragraph{FUNSD}

FUNSD~\citep{Jaume_2019} is a dataset for form understanding in noisy scanned documents. It contains 199 real, fully annotated, scanned forms where 9,707 semantic entities are annotated above 31,485 words. The 199 samples are split into 149 for training and 50 for testing. The official OCR annotation is directly used with the layout information. The FUNSD dataset is suitable for a variety of tasks, where we focus on semantic entity labeling in this paper. Specifically, the task is assigning to each word a semantic entity label from a set of four predefined categories: question, answer, header, or other. The entity-level F1 score is used as the evaluation metric.

\paragraph{CORD}

We also evaluate our model on the receipt key information extraction dataset, i.e. the public available subset of CORD~\citep{park2019cord}. The dataset includes 800 receipts for the training set, 100 for the validation set, and 100 for the test set. A photo and a list of OCR annotations are equipped for each receipt. An ROI that encompasses the area of receipt region is provided along with each photo because there can be irrelevant things in the background. We only use the ROI as input instead of the raw photo. The dataset defines 30 fields under 4 categories and the task aims to label each word to the right field. The evaluation metric is entity-level F1. We use the official OCR annotations.

\paragraph{SROIE}

The SROIE dataset (Task 3)~\citep{8977955} aims to extract information from scanned receipts. There are 626 samples for training and 347 samples for testing in the dataset. The task is to extract values from each receipt of up to four predefined keys: company, date, address, or total. The evaluation metric is entity-level F1. We use the official OCR annotations and results on the test set are provided by the official evaluation site.

\paragraph{Kleister-NDA}

Kleister-NDA~\citep{graliski2020kleister} contains non-disclosure agreements collected from the EDGAR database, including 254 documents for training, 83 documents for validation, and 203 documents for testing. This task is defined to extract the values of four fixed keys. We get the entity-level F1 score from the official evaluation tools.\footnote{\url{https://gitlab.com/filipg/geval}} Words and bounding boxes are extracted from the raw PDF file. We use heuristics to locate entity spans because the normalized standard answers may not appear in the utterance. As the labeled answers are normalized into a canonical form, we apply post-processing heuristics to convert the extracted date information into the ``YYYY-MM-DD'' format, and company names into the abbreviations such as ``LLC'' and ``Inc.''.

\paragraph{RVL-CDIP}

RVL-CDIP~\citep{harley2015icdar} consists of 400,000 grayscale images, with 8:1:1 for the training set, validation set, and test set. A multi-class single-label classification task is defined on RVL-CDIP. The images are categorized into 16 classes, with 25,000 images per class. The evaluation metric is the overall classification accuracy. Text and layout information is extracted by Microsoft OCR.

\paragraph{DocVQA}

As a VQA dataset on the document understanding field, DocVQA~\citep{Mathew_2021_WACV} consists of 50,000 questions defined on over 12,000 pages from a variety of documents. Pages are split into the training set, validation set, and test set with a ratio of about 8:1:1. The dataset is organized as a set of triples $\langle$page image, questions, answers$\rangle$. Thus, we use Microsoft Read API to extract text and bounding boxes from images. Heuristics are used to find given answers in the extracted text. The task is evaluated using an edit distance based metric ANLS (aka average normalized Levenshtein similarity). Given that human performance is about 98\% ANLS on the test set, it is reasonable to assume that the found ground truth which reaches over 97\% ANLS on training and validation sets is good enough to train a model. Results on the test set are provided by the official evaluation site.

\section{Model Training Details}

\paragraph{Pre-training}

We pre-train LayoutLMv2 models using Adam optimizer \citep{kingma2017adam,loshchilov2018decoupled}, with the learning rate of $2 \times 10^{-5}$, weight decay of $1 \times 10^{-2}$,and $(\beta_1, \beta_2) = (0.9, 0.999)$. The learning rate is linearly warmed up over the first $10\%$ steps then linearly decayed. $\textrm{LayoutLMv2}_{\rm BASE}$ is trained with a batch size of $64$ for $5$ epochs, and $\textrm{LayoutLMv2}_{\rm LARGE}$ is trained with a batch size of $2048$ for $20$ epochs on the IIT-CDIP dataset.

\paragraph{Fine-tuning for Visual Question Answering}

We treat the DocVQA as an extractive QA task and build a token-level classifier on top of the text part of LayoutLMv2 output representations. Question tokens, context tokens and visual tokens are assigned to segment {\tt[A]}, {\tt[B]} and {\tt[C]}, respectively. The maximum sequence length is set to $L = 384$.

\paragraph{Fine-tuning for Document Image Classification}

This task depends on high-level visual information, thereby we leverage the image features explicitly in the fine-tuning stage. We pool the visual embeddings into a global pre-encoder feature, and pool the visual part of LayoutLMv2 output representations into a global post-encoder feature. The pre and post-encoder features along with the {\tt[CLS]} output feature are concatenated and fed into the final classification layer.

\paragraph{Fine-tuning for Sequential Labeling}

We formalize FUNSD, SROIE, CORD, and Kleister-NDA as the sequential labeling tasks. To fine-tune LayoutLMv2 models on these tasks, we build a token-level classification layer above the text part of the output representations to predict the BIO tags for each entity field.

\begin{table*}[ht]
	\centering
	\small
	\begin{tabular}{lccc}
		\toprule
		\multicolumn{1}{c}{\bf Model} & \bf Precision & \bf Recall & \bf F1  \\\midrule
		$\textrm{BERT}_{\rm BASE}$  & 0.5469 & 0.6710 & 0.6026  \\
		$\textrm{UniLMv2}_{\rm BASE}$  & 0.6561 & 0.7254 & 0.6890   \\
		$\textrm{BERT}_{\rm LARGE}$   & 0.6113 & 0.7085 & 0.6563  \\
		$\textrm{UniLMv2}_{\rm LARGE}$  & 0.6903 & 0.7649 & 0.7257  \\\midrule
		$\textrm{LayoutLM}_{\rm BASE}$ & 0.7597 & 0.8155 & 0.7866  \\
		$\textrm{LayoutLM}_{\rm LARGE}$ & 0.7596 & 0.8219 & 0.7895  \\\midrule
		$\textrm{LayoutLMv2}_{\rm BASE}$ & 0.8029 & 0.8539 & 0.8276  \\
		$\textrm{LayoutLMv2}_{\rm LARGE}$ & \bf 0.8324 & \bf 0.8519 & \bf 0.8420  \\\midrule
		\midrule
		%   BROS\tablefootnote{\url{ https://openreview.net/pdf?id=punMXQEsPr0}} & 0.8056 & 0.8188 & 0.8121 & - \\
		BROS ~\citep{hong2021bros} & 0.8056 & 0.8188 & 0.8121  \\
		\bottomrule
	\end{tabular}
	\caption{Model accuracy (entity-level Precision, Recall, F1) on the FUNSD dataset}
	\label{tab:funsd}
\end{table*}

\begin{table*}[ht]
	\centering
	\small
	\begin{tabular}{lccc}
		\toprule
		\multicolumn{1}{c}{\bf Model} & \bf Precision & \bf Recall & \bf F1 \\
		\midrule
		$\textrm{BERT}_{\rm BASE}$ & 0.8833 & 0.9107 & 0.8968 \\
		$\textrm{UniLMv2}_{\rm BASE}$ & 0.8987 & 0.9198 & 0.9092  \\
		$\textrm{BERT}_{\rm LARGE}$ & 0.8886 & 0.9168 & 0.9025 \\
		$\textrm{UniLMv2}_{\rm LARGE}$ & 0.9123 & 0.9289 & 0.9205  \\
		\midrule
		$\textrm{LayoutLM}_{\rm BASE}$ & 0.9437 & 0.9508 & 0.9472 \\
		$\textrm{LayoutLM}_{\rm LARGE}$ & 0.9432 & 0.9554 & 0.9493 \\
		\midrule
		$\textrm{LayoutLMv2}_{\rm BASE}$ & 0.9453 & 0.9539 & 0.9495 \\
		$\textrm{LayoutLMv2}_{\rm LARGE}$ & \bf 0.9565 & \bf 0.9637 & \bf 0.9601  \\
		\midrule\midrule
		SPADE~\citep{hwang2020spatial} & - & - & 0.9150  \\
		BROS~\citep{hong2021bros} & 0.9558 & 0.9514 & 0.9536  \\
		\bottomrule
	\end{tabular}
	\caption{Model accuracy (entity-level Precision, Recall, F1) on the CORD dataset}
	\label{tab:cord}
\end{table*}

\begin{table*}[ht]
	\centering
	\small
	\begin{tabular}{lccc}
		\toprule
		\multicolumn{1}{c}{\bf Model} & \bf Precision & \bf Recall & \bf F1   \\\midrule
		$\textrm{BERT}_{\rm BASE}$  & 0.9099 & 0.9099 & 0.9099  \\
		$\textrm{UniLMv2}_{\rm BASE}$  & 0.9459 & 0.9459 & 0.9459   \\
		$\textrm{BERT}_{\rm LARGE}$   & 0.9200 & 0.9200 & 0.9200  \\
		$\textrm{UniLMv2}_{\rm LARGE}$  & 0.9488 & 0.9488 & 0.9488  \\\midrule
		$\textrm{LayoutLM}_{\rm BASE}$ & 0.9438 & 0.9438 & 0.9438  \\
		$\textrm{LayoutLM}_{\rm LARGE}$ & 0.9524 & 0.9524 & 0.9524  \\\midrule
		$\textrm{LayoutLMv2}_{\rm BASE}$ & 0.9625 & 0.9625 & 0.9625  \\
		$\textrm{LayoutLMv2}_{\rm LARGE}$ & 0.9661 &  0.9661 & 0.9661  \\
		$\textrm{LayoutLMv2}_{\rm LARGE}$ (Excluding OCR mismatch) & \bf 0.9904 & \bf 0.9661 & \bf 0.9781  \\
		\midrule\midrule
		BROS~\citep{hong2021bros} & 0.9493 & 0.9603 & 0.9548  \\
		PICK~\citep{yu2020pick} & 0.9679 & 0.9546  & 0.9612 \\
		TRIE~\citep{zhang2020trie} & - & - & 0.9618 \\
		Top-1 on SROIE Leaderboard (Excluding OCR mismatch) & 0.9889 & 0.9647  & 0.9767  \\
		\bottomrule
	\end{tabular}
	\caption{Model accuracy (entity-level Precision, Recall, F1) on the SROIE dataset (until 2020-12-24)}
	\label{tab:sroie}
\end{table*}

\begin{table*}[ht]
	\centering
	\small
	\begin{tabular}{lc}
		\toprule
		\multicolumn{1}{c}{\bf Model} & \bf F1   \\\midrule
		$\textrm{BERT}_{\rm BASE}$ & 0.779  \\
		$\textrm{UniLMv2}_{\rm BASE}$ & 0.795   \\
		$\textrm{BERT}_{\rm LARGE}$ & 0.791  \\
		$\textrm{UniLMv2}_{\rm LARGE}$ & 0.818  \\\midrule
		$\textrm{LayoutLM}_{\rm BASE}$ & 0.827  \\
		$\textrm{LayoutLM}_{\rm LARGE}$ & 0.834  \\\midrule
		$\textrm{LayoutLMv2}_{\rm BASE}$ & 0.833  \\
		$\textrm{LayoutLMv2}_{\rm LARGE}$ & \bf 0.852  \\\midrule\midrule
		$\textrm{RoBERTa}_{\rm BASE}$ in~\citep{graliski2020kleister} & 0.793  \\
		\bottomrule
	\end{tabular}
	\caption{Model accuracy (entity-level F1) on the validation set of the Kleister-NDA dataset using the official evaluation toolkit}
	\label{tab:kleister-nda}
\end{table*}

\section{Detailed Experiment Results}

Tables list per-task detailed results for the four entity extraction tasks, with Table \ref{tab:funsd} for FUNSD, Table \ref{tab:cord} for CORD, Table \ref{tab:sroie} for SROIE, and Table \ref{tab:kleister-nda} for Kleister-NDA.

\end{document}